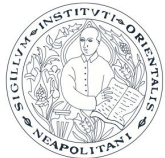
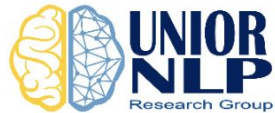

# *#LaCulturaNonSiFerma*

Report sull'uso e la diffusione degli #hashtag delle istituzioni culturali italiane durante il periodo di lockdown

**Autori**: Carola Carlino, Gennaro Nolano, Maria Pia di Buono, Johanna Monti - Università di Napoli "L'Orientale" - UNIOR NLP Research Group
**Data**: 20/05/2020
**Versione**: 1.0

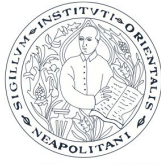
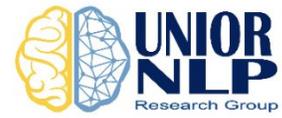

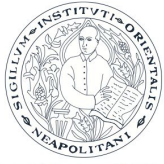
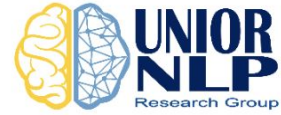

# 1. Introduzione

Durante la pandemia di COVID-19 del 2019-2020, molti stati hanno fatto ricorso ad una politica di lockdown, in alcuni casi ancora in corso, che ha interessato milioni di cittadini, istituzioni e aziende per il contenimento e contrasto del diffondersi del virus. L'Italia, insieme alla Cina, è stata tra i primi paesi ad applicare tale politica di restrizioni ad ampio raggio che ha, ovviamente, interessato anche istituzioni e luoghi di cultura.

In tale periodo, diverse iniziative comunicative a vari livelli sono state intraprese per aumentare la consapevolezza dei cittadini in riferimento alla necessità di modificare temporaneamente comportamenti e abitudini di vita e offrire supporto e incoraggiamento per affrontare il nuovo scenario.

In particolare, le istituzioni culturali, quali musei e gallerie d'arte, hanno dato vita a un'ampia gamma di azioni, utilizzando le principali piattaforme social (Facebook, Twitter, Instagram) e richiedendo la partecipazione attiva degli utenti a tali iniziative. Il coinvolgimento degli utenti, che, in fase di lockdown, poteva avvenire solo tramite una forma mediata, sembra perseguire due obiettivi principali: (i) mantenere attivo il canale comunicativo con il pubblico; (ii) garantire l'accesso ai contenuti culturali non fruibili di persona a causa delle limitazioni dovute alla politica di restrizione degli spostamenti fisici per tutti i cittadini.

L'uso degli strumenti social si è rivelato fondamentale per assolvere la funzione culturale e la funzione comunicativa, che, come ricorda Antinucci (2004:10), "sono inscindibili, giacché l'una presuppone l'altra". Privati della possibilità di dialogare con il pubblico in maniera diretta e di comunicare i loro contenuti nel contesto tradizionale di fruizione, le istituzioni hanno sfruttato pienamente le potenzialità della nuova modalità comunicativa.

Uno dei fenomeni, immediatamente evidente, nella maggior parte delle interazioni promosse da istituzioni culturali attraverso tutti i social media, è l'utilizzo di una serie di #hashtag (#) ad accompagnare il contenuto visivo e testuale proposto. Gli #hashtag, parole chiave assegnate alle informazioni, nate per descrivere i contenuti su Twitter e guidarne la stessa ricerca, rappresentano un elemento centrale per organizzare le informazioni, in quanto contribuiscono a raggruppare le discussioni relative ad argomenti o eventi specifici (Small, 2011).

Alcuni studi in altri settori, in particolare nei settori del marketing e della comunicazione, ma anche di *Business Intelligence*, riconoscono il ruolo degli #hashtag nel contribuire all'innovazione delle *knowledge organizations* in un contesto, come quello attuale, in cui proliferano contenuti generati dagli utenti (Chang, 2010). Infatti, secondo lo studio proposto da Chang (2010), la teoria della *Diffusion of Innovation* può contribuire a esaminare la diffusione nell'uso e adozione degli #hashtag all'interno di un sistema social. La necessità di organizzare le informazioni e i contenuti intorno ad argomenti specifici, in modo da facilitarne il successivo recupero, ha causato una ampia adozione degli #hashtag. Tale adozione ha dato vita a veri e propri archivi di informazioni, risultato di uno sforzo collettivo, cui partecipano attivamente tutti coloro che contribuiscono ad aggregare i contenuti nello stesso flusso identificato da uno specifico #hashtag.

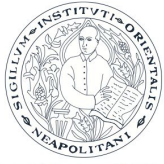
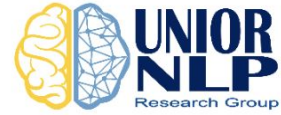

A dimostrazione dell'importanza degli #hashtag nell'aggregare contenuti e utenti, nel 2015 Twitter introduce l'#hashtag *#culturalheritage,* per rispondere all'interesse dimostrato degli utenti verso i contenuti e gli eventi culturali. L'opportunità offerta dalla connessione tra #hashtag e contenuti culturali è stata analizzata in studi recenti, anche se con una limitata attenzione al potenziale e all'efficacia degli impulsi e attività social per supportare le istituzioni nella promozione dei beni culturali (Chianese et al., 2016).

L'uso degli #hashtag si è successivamente ampliato travalicando i confini di Twitter e diventando un elemento portante nella costruzione di comunità virtuali che si identificano nel messaggio e intorno al quale costruiscono relazioni, migliorando la comunicazione tra i membri che aderiscono alla comunità (Mulyadi and Fitriana, 2018).

Marulli et al. (2015) propongono un approccio basato su ontologie per valutare l'interesse e la sensibilità degli utenti ai contenuti dei beni culturali attraverso le loro attività sui social network, in particolare Twitter. I risultati di tale indagine quantitativa, che combina il trattamento automatico del linguaggio (TAL), le tecnologie semantiche e l'analisi temporale e geo-referenziata, misurano l'atteggiamento degli utenti rispetto agli oggetti dei beni culturali, la densità geografica dei luoghi in cui fruire contenuti culturali e la prossimità temporale ad eventi legati ai beni culturali.

L'analisi dell'uso e della diffusione degli #hashtag, quindi, rappresenta, alla luce delle numerose variabili che possono essere considerate, una interessante sorgente di informazioni sui comportamenti delle istituzioni e degli utenti che interagiscono con queste e con i contenuti culturali in generale.

Il presente report si basa su una prima indagine degli #hashtag prodotti e utilizzati dalle istituzioni culturali italiane durante il periodo di lockdown nell'arco temporale che va dal 08/03/2020, data in cui è stato emanato il Decreto del Presidente del Consiglio dei Ministri (DPCM 8 marzo 2020) che corrisponde a quella che è stata denominata Fase 1 e che ha sancito l'inizio delle restrizioni dovute alle misure per il contenimento e il contrasto del Covid-19 sull'intero territorio nazionale, al 04/05/2020, data in cui il DPCM 26 Aprile 2020 stabiliva l'allentamento di tali misure dando inizio alla cosiddetta Fase 2[1].

Il particolare periodo preso in considerazione rappresenta uno scenario interessante di ricerca per analizzare come le istituzioni culturali hanno risposto ad un evento che, di fatto, mutava le condizioni di comunicazione in cui operare e come gli utenti hanno reagito a questa nuova forma in cui sono stati proposti dei contenuti culturali.

---

[1] È importante sottolineare che il DPCM 26 Aprile 2020 non sancisce la riapertura dei luoghi della cultura al 4 maggio, bensì al 18 maggio. Tuttavia, si è scelto di concentrare l'analisi e la raccolta dei dati in riferimento al primo periodo di lockdown in quanto più restrittivo per tutte le attività della popolazione italiana e che, quindi, può essere considerato maggiormente interessante per valutare le reazioni degli utenti alle iniziative delle istituzioni culturali.

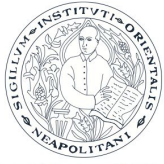 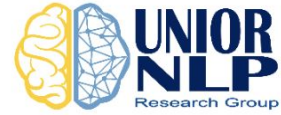

Lo studio è stata condotto tenendo conto di tre livelli di analisi: (i) il modo in cui le istituzioni culturali hanno curato la comunicazione attraverso i social nel periodo definito Fase 1 per confermare o rafforzare la loro presenza, proponendo nuove iniziative e contenuti agli utenti; (ii) il grado di partecipazione da parte del pubblico alle iniziative; (iii) la dimensione temporale che ha causato la sospensione improvvisa delle attività culturali in presenza, modificando drasticamente il comportamento del pubblico di tali istituzioni e azzerando di fatto la fruizione diretta dei contenuti culturali proposti dai luoghi di cultura.

## 2. #Hashtag e iniziative culturali durante il lockdown

Durante il periodo di lockdown, numerose iniziative di divulgazione dei contenuti culturali attraverso i social network sono state caratterizzate dall'uso di #hashtag. Tale uso, contribuendo all'aggregazione delle informazioni, ha incentivato la creazione di un luogo di cultura virtuale in cui musei e istituzioni hanno concentrato le loro attività di comunicazione con il loro pubblico tradizionale e con potenziali nuovi utenti.

Dinanzi all'ampio ventaglio di proposte che i musei hanno lanciato, l'International Council Of Museums[2] (ICOM), la principale organizzazione internazionale che rappresenta i musei e i suoi professionisti, già pochi giorni dopo la diffusione di tali iniziative, ha stilato un elenco contenente i casi studio che sono sembrati essere più stimolanti e meglio pensati per raggiungere il pubblico da remoto[3]. Tra questi rientrano coloro che hanno digitalizzato il contenuto delle proprie collezioni, come ad esempio lo Smithsonian, l'istituto di ricerca fondato a Washington nel 1846, che ha pubblicato su una piattaforma accessibile a tutti 2,8 milioni di immagini ad alta risoluzione[4], e coloro che hanno incrementato l'utilizzo dei canali social, Facebook primo tra tutti, seguito da Instagram, Twitter e altri[5], offrendo agli utenti tour in diretta tra i propri capolavori, spesso illustrati dai direttori dei musei, come ad esempio la passeggiata virtuale offerta dal Los Angeles County Museum of Art[6] (LACMA).

Se Instagram è stato usato come vetrina attraverso la quale artisti e professionisti del settore hanno intrattenuto momenti conversevoli con un pubblico di interessati utilizzando la formula della "diretta", Twitter, che permette un approccio meno interattivo in tal senso, è stato adottato da alcuni musei come strumento per coinvolgere il proprio pubblico in modo divertente, proponendo quiz e giochi creati sulle opere delle proprie collezioni, quali i puzzle e le parole crociate elaborati dal Getty Museum di Los Angeles[7] e il Museum of Fine Arts della città di Saint Petersburg in Florida[8].

---

[2] http://www.icom-italia.org/
[3] https://icom.museum/en/news/how-to-reach-and-engage-your-public-remotely/
[4] https://www.si.edu/openaccess
[5] Sull'uso dei social network più attrattivi per il pubblico si veda il riferimento ai risultati del report finale pubblicato dalla *Network of European Museum Organisations* (NEMO) il 12/05/2020 https://www.ne-mo.org/fileadmin/Dateien/public/NEMO_documents/NEMO_COVID19_Report_12.05.2020.pdf
[6] https://www.facebook.com/LACMA/videos/vl.1088362991231549/10155304471391566/?type=1

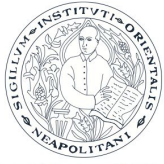 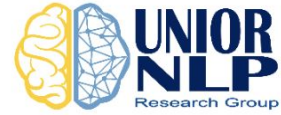

Zoomando sull'Italia, ci si rende conto che anche qui le istituzioni nazionali - musei, gallerie, fondazioni - non hanno atteso a far avvertire la propria presenza e hanno messo in atto tutte le risorse a disposizione per restare in contatto con il proprio pubblico[9], garantendo l'accesso al nuovo luogo di cultura virtuale attraverso l'uso degli #hashtag.

Il presente studio si basa sull'analisi di un set di #hashtag utilizzati per rispondere alla situazione inaspettata di lockdown che ha creato un gap comunicativo improvviso con gli utenti. In generale, l'utilizzo degli #hashtag in tale contesto ha assunto due compiti: mantenere attivo un canale di comunicazione con il pubblico e offrire modalità di fruizione alternative dei contenuti culturali di cui tali istituzioni sono depositarie.

# 3. Raccolta e Analisi dei Dati

L'indagine ha preso in considerazione lo specifico spazio virtuale creato dagli #hashtag che le istituzioni culturali hanno utilizzato nella promozione e nella comunicazione durante la Fase 1, in particolare sul social network Twitter. Le specifiche caratteristiche di questa piattaforma hanno fatto sì che la sua adozione da parte delle istituzioni culturali italiane fosse graduale, se si considerano i dati rilevati dall'Osservatorio Innovazione Digitale nei Beni e Attività Culturali in seguito a un'analisi effettuata su un campione di musei in relazione all'attività da loro svolta nel 2016[10]. Da quegli anni ad oggi, la percentuale dei musei italiani che usano Twitter è passata dal 31% al 33% nel biennio successivo (2017-18)[11].

Il periodo di lockdown ha drasticamente incrementato le attività e la presenza online delle istituzioni, come indicato dal report NEMO, in cui sono stati analizzati differenti dati relativi alla chiusura forzata. In particolare, il report evidenzia che, dopo tre settimane di chiusura al pubblico, circa l'80% dei musei ha incrementato la propria attività online, come reazione alla generale visibilità in crescita dei beni culturali fruiti in questa modalità. In riferimento a Twitter, si rivolge l'attenzione non tanto a quanti musei hanno iniziato a usarlo, ma piuttosto a come hanno iniziato a usufruirne, poiché, se prima della pandemia la piattaforma social era considerata soltanto uno dei tanti mezzi di comunicazione, con un minimo obiettivo di engagement del pubblico, a causa delle misure di *social distancing* sembra essere diventato una risorsa irrinunciabile. Tuttavia, l'impiego di questa piattaforma richiede una buona capacità di sintesi e un linguaggio efficacemente adoperato per raccontare eventi e fornire informazioni ad esso correlate, come nel caso di una mostra o di qualunque iniziativa lanciata da un museo (Mandarano, 2019:102). D'altro canto, Marulli et al. (2015) ricordano come il settore dei beni culturali possa trarre vantaggio da queste tecnologie della comunicazione e dell'informazione, sviluppando nuove strategie per coinvolgere e attrarre un numero maggiore di utenti. Il particolare contesto, creatosi per l'evento pandemico, ha causato la necessità di modificare il modo in cui le istituzioni culturali sfruttano questi canali, ampliando i fini per i quali le stesse istituzioni hanno iniziato a usare tali tecnologie, ovvero attrarre visitatori alle mostre o promuovere una serie di servizi collegati.

---

[7] https://medium.com/@AkronArtMuseum/musegames-round-two-60b71939e030
[8] https://www.jigsawplanet.com/?rc=play&pid=01e489a945b2
[9] https://www.artribune.com/arti-visive/arte-contemporanea/2020/03/iorestoacasa-musei-covid-19-contenuti-online-guida/78/
[10] https://www.osservatori.net/it_it/osservatori/comunicati-stampa/il-52-dei-musei-italiani-e-social-ma-i-servizi-digitali-per-la-fruizione-delle-opere-sono-limitati
[11] https://www.statista.com/statistics/553443/share-of-museums-using-on-social-media-by-platform-italy/

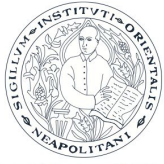 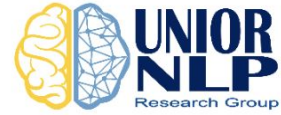

## 3.1 Raccolta dei dati e creazione del dataset

La raccolta dei dati in forma di tweet è stata effettuata a partire da un set di #hashtag predominanti nei post delle principali istituzioni culturali e del MiBACT sulle piattaforme social, durante il momento iniziale della cosiddetta Fase 1. La prima selezione manuale ha incluso un totale di 33 #hashtag, fra quelli riportati dalla piattaforma Twitter tra i *trend topics* in tale periodo. Sulla base di questa selezione iniziale, è stato creato un primo dataset includendo i tweet pubblicati nell'arco di tempo che va dal 08/03/2020 al 04/05/2020, che presentassero uno o più degli #hashtag individuati (Annex - Tabella 1). In totale, questo dataset contiene 23716 tweet, che sono stati raccolti attraverso Python 3.6 e la librerie GetOldTweets[12].

Il set iniziale di #hashtag, ordinato in base al numero di retweet per ciascuno, è stato successivamente suddiviso in base a un criterio temporale di creazione degli aggregatori stessi, individuando due macro-categorie: quelli esistenti prima della diffusione del virus COVID-19 e dell'entrata in vigore delle relative misure di restrizione, e quelli che sono stati creati *ad hoc* durante periodo di lockdown.

Per quanto riguarda gli #hashtag del primo gruppo (riportati in grigio in Tabella 1), questi compaiono sulla homepage di Twitter già in un momento precedente a quello compreso tra la fine del 2019 e l'inizio del 2020, ma al fine di incentivare il pubblico costretto a stare in casa a fruire dei contenuti culturali, essi sono stati riutilizzati dai musei per sostenere le proprie iniziative.

È il caso, ad esempio, dell'hashtag *#ABCBarberiniCorsini*, che riassume l'idea elaborata dallo staff della Galleria Barberini Corsini di Roma di diffondere la conoscenza delle opere della propria collezione. La proposta è stata lanciata il 16 ottobre 2019 ed è stata nuovamente sottoposta all'attenzione del pubblico l'11 marzo 2020, vale a dire due giorni dopo la chiusura dei musei sancita dall'emanazione del DPCM 8 marzo 2020.

Tra gli #hashtag più generici, cioè quelli che non fanno riferimento a una singola iniziativa, ma che rimandano a delle macrocategorie o macrotemi, si nota una corposa interazione dell'hashtag *#museitaliani*, che sui social network è presente anche nella variante grafica *#museiitaliani*. Sulle diverse piattaforme social è l'#hashtag ufficiale adottato dalla Direzione generale Musei[13], organo funzionario del MiBACT, ed è l'aggregatore con il quale si vogliono rendere note le opere, i capolavori, il luoghi della cultura del Sistema Museale Nazionale italiano. Proprio per la *mission* per cui nasce questa etichetta, si può affermare che essa esiste da prima del filone Musei e COVID, ma di sicuro ha contribuito non poco a rafforzare la presenza anche dei musei italiani, per l'appunto, come interlocutori di un dialogo che, senza voce, coinvolge utenti e istituzioni di tutto il mondo.

---

[12] https://github.com/Jefferson-Henrique/GetOldTweets-python/blob/master/README.md

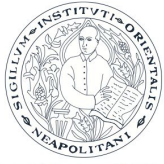
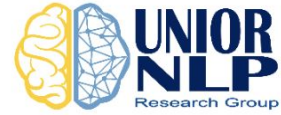

Diversa è l'opinione per *#artchallenge*, che, sebbene sia stato utilizzato in molti testi brevi, presenta ben pochi riferimenti al contesto pandemico. Dal punto di vista geografico, il suo uso si colloca nei paesi oltreoceano e oltremanica; sui casi di applicazione, invece, risulta essere adoperato a grandi linee da utenti che si occupano di grafica, di disegno e che hanno tra i propri interessi la cultura dei Manga.

Nella stessa [Tabella 1](), sono riportati anche gli #hashtag che sono stati creati dai musei per far sentire fortemente la propria presenza sui social e per fare in modo che il pubblico ne fosse consapevole, nonostante la momentanea chiusura forzata delle sale di esposizione.

Al fine di evitare rumore nei dati e focalizzarci su un set ridotto di #hashtag, nella analisi successive abbiamo deciso di non considerare gli #hashtag con numero totale di tweet nel periodo di interesse inferiore a 1000 e gli #hashtag che, pur avendo un numero di tweet relativi abbastanza elevato, sono stati usati prevalentemente da organizzazioni straniere o in contesti non strettamente connessi ai beni culturali (*#cultureinquarantine*, *#artchallenge* in [Tabella 1]()). Il numero totale di #hashtag selezionati è sei, di cui alcuni pre-esistenti al periodo di lockdown, come *#museitaliani*, ma riutilizzati in occasione delle nuove attività del periodo pandemico.Tale selezione ha portato ad una cernita dei tweet presenti nel dataset iniziale, prendendo in esame un totale di 15988 tweet prodotti utilizzando almeno uno dei sei #hashtag di interesse (Annex - [Tabella 2]()).

Nella [Tabella 2]() è possibile avere un quadro del numero di interazioni, in termini di dati disaggregati in tweet e retweet, per ognuno dei sei #hashtag presi in considerazione.

Sulla base del numero di interazioni generate, un discorso e un'analisi a parte merita *#ArTyouReady*, #hashtag di nuova creazione, che con 16237 retweet nel periodo di Fase 1, si candida a ricoprire il ruolo di aggregatore tematico più diffuso e noto.

Infine, attraverso la libreria Tweepy[14] sono state poi raccolte informazioni aggiuntive sugli utenti, in particolare: quanti e quali utenti fossero verificati e l'area geografica di appartenenza. Dagli utenti verificati, infine, sono stati separati gli account appartenenti a istituzioni e organizzazioni culturali italiane.

---

[13] http://musei.beniculturali.it/
[14] https://www.tweepy.org/

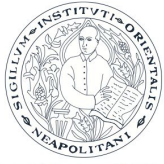
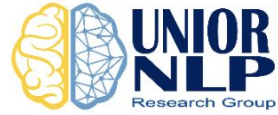

## 3.2 Analisi preliminare

In questa prima fase di analisi i dati sono stati aggregati utilizzando diverse variabili, prima tra tutte quella temporale. Infatti, l'andamento di ognuno dei sei #hashtag scelti in termini di interazioni nell'arco temporale in considerazione è indice sia della capacità delle istituzioni di rispondere alla situazione mutata sia del successo dell'azione comunicativa delle istituzioni.

**Creazione e uso degli #hashtag**. Il grafico in [Figura 1](#) riporta l'andamento degli #hashtag che sono stati analizzati all'interno del periodo di riferimento. Come si evince dal picco in esso rappresentato, si registra un incremento nell'uso di quasi tutti gli #hashtag, ad eccezione di *#cultureinquarantine* e *#laculturaincasa*, in corrispondenza del giorno 29/03. Tale data corrisponde al lancio dell'iniziativa *#ArTyouReady*.

Questo #hashtag e l'iniziativa ad esso collegata hanno avuto una risonanza molto ampia, come dimostrato non solo dal picco nel numero di interazioni generate ma anche dal fatto che a parlarne sono stati i tg nazionali[15] e alcune testate giornalistiche internazionali[16].

La proposta promossa dal Ministero per i beni e le attività culturali e per il turismo[17] (MiBACT) è stata inizialmente concepita come un flashmob digitale dedicato alla fotografia ed è stato richiesto agli utenti di condividere gli scatti che essi avevano precedentemente effettuato all'interno dei musei; un tentativo per mantenere vivo il patrimonio culturale racchiuso all'interno dei musei resi necessariamente deserti, e ciò che la contraddistingue è la combinazione di #hashtag che di volta in volta si accompagnano a quello principale, ovvero *#ArTyouReady*, come etichetta che segnala il significato delle proposte che vengono avanzate.

La prima edizione del 29 marzo 2020 prevede una riflessione sulla chiusura dei musei che viene resa con la formula *#emptymuseum* e a questa ne sono seguite altre che per tutto il mese di aprile, ogni domenica, hanno coinvolto gli utenti in modo diverso.

Il 5 aprile sul sito web del MiBACT e sui propri canali social viene lanciato l'#hashtag *#ArTyouReady* insieme al *#GrandVirtualTour*, con lo scopo di ottenere da parte degli utenti una ricostruzione di un viaggio digitale lungo tutta la Penisola postando le foto che essi hanno scattato nei luoghi della cultura.

La proposta calendarizzata il 12 aprile arricchisce l'iniziativa originaria di un aspetto più creativo, perché con l'*#artetisomiglia*, storica campagna istituzionale del MiBACT risalente al gennaio 2017, gli utenti si divertono a postare sui social foto ritraenti sé stessi mentre imitano un'opera d'arte con ciò che hanno in casa. L'evento, dalla portata internazionale, è stato replicato anche dal Getty Museum di Los Angeles, che si è lasciato ispirare dal Rijksmuseum di Amsterdam e da un account instagram denominato **#*tussenqunstenquarantine*[18]** (Tra Arte e Quarantena).

---

[15] https://www.youtube.com/watch?v=A5AHwIlcGZw
[16] https://www.huffingtonpost.it/entry/art-you-ready_it_5e7faaf3c5b6cb9dc1a1697b
[17] https://www.beniculturali.it/mibac/export/MiBAC/sito-MiBAC/Contenuti/MibacUnif/Comunicati/visualizza_asset.html_2101868381.html

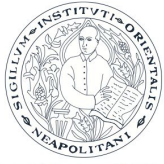
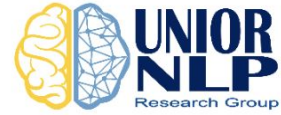

I protagonisti del quarto appuntamento di *#ArTyouReady*, previsto per il giorno 19 aprile, sono i giardini, il mare, le colline, i borghi, le montagne e i tramonti che rivivono negli scatti fatti dagli utenti che apprezzano il *#paesaggioitaliano*; è proprio questo l'#hashtag usato per la penultima edizione del flashmob fotografico.

In occasione dell'ultimo incontro, avvenuto il 26 aprile, il MiBACT in collaborazione con Office de Tourisme Italien - ENIT e Touring Club Italiano, sviluppando l'idea precedente del paesaggio italiano, conclude il tema del viaggio che, dalla seconda alla quarta edizione è passato dall'essere virtuale all'essere fisico; con l'#hashtag *#viaggioinitalia* si condividono sui social foto di antiche mappe che ricostruiscono la morfologia del territorio italiano negli anni.

**Diffusione degli #hashtag.** La Figura 2 riporta l'andamento degli #hashtag usati nei tweet nel mese di marzo. Come si può vedere, dopo un leggero incremento registrato in corrispondenza del 13/03, la linea rimane piatta fino ad arrivare al giorno 28/03, data in cui inizia a crescere per raggiungere il picco massimo il 29/03, e decrescere nei giorni successivi, con una leggera variazione nei giorni 04/04, 05/04, 06/04, dovuta al rilancio dell'iniziativa *#ArTyouReady* sui canali social.

La Figura 3 riporta l'andamento degli #hashtag usati nei tweet nel mese di aprile. Tra quelli presenti, si può notare un trend costante per l'#hashtag *#emptymuseum*, con un leggero incremento il giorno 23/04. L#'hashtag *#museichiusimuseiaperti* presenta una crescita più o meno costante, con un sensibile incremento in corrispondenza del periodo compreso tra il 29/04 e il 01/05. I momenti di crescita che riguardano l'#hashtag *#ArTyouReady* corrispondo ai giorni nei quali l'iniziativa è stata riproposta dalle istituzioni (19/04, 26/04 e 03/05) e a quelli immediatamente precedenti, durante i quali è stata "pubblicizzata". Per tutte le altre etichette, si nota un considerevole incremento a cavallo tra la fine del mese di aprile e l'inizio del mese di maggio, nel periodo circoscritto ai giorni 22/04 e 04/05. L'incremento registrato in questo lasso temporale è probabilmente dovuto alla corrispondenza con altre attività che hanno avuto luogo nei giorni di maggior interesse: Giornata mondiale del libro (23/05/2020), Giornata Internazionale della Danza (29/04) e Festa dei lavoratori (01/05/2020). Gli #hashtag che sono associati a queste iniziative, rispettivamente *#giornatamondialedellibro*, *#giornatamondialedelladanza* e *#Festadeilavoratori*, sono utilizzati dagli utenti di twitter insieme all'#hashtag più generale *#laculturanonsiferma*, come evidenziato dai risultati di un ricerca manuale effettuata su Twitter.

La Figura 4 riporta l'andamento dei retweet per #hashtag in riferimento al mese di marzo. Si può notare un incremento delle attività in corrispondenza dei giorni 28 - 30 marzo, e un più leggero aumento in riferimento ai primi giorni del mese di aprile.

---

[18] https://www.instagram.com/tussenkunstenquarantaine/

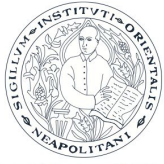 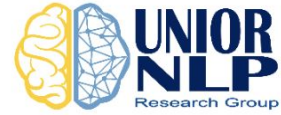

La Figura 5 riporta l'andamento dei retweet per #hashtag in riferimento all'intero mese di aprile e ai primi quattro giorni del mese di maggio. Il trend di crescita delle interazioni è più o meno costante per tutti gli #hashtag, fatta eccezione per l'#hashtag *#ArTyouReady* che è stato maggiormente ritwittato dagli utenti nei giorni 17, 18, 19, 20, 25, 26 e 27 aprile, e nei primi giorni di maggio.

In Tabella 3 è riportato il numero di utenti unici che hanno risposto alle iniziative scrivendo dei tweet nei quali compare uno o più #hashtag tra quelli selezionati. Per 'utenti unici' si intendono quegli utenti che hanno utilizzato l'etichetta almeno una volta nella formulazione dei propri tweet, e sono stati, quindi, registrati una sola volta nel conteggio automatico. Il totale riportato per ogni hashtag comprende gli utenti semplici e le istituzioni ufficiali, sulla cui distinzione si rimanda al paragrafo successivo. I dati rivelano un ampio uso dell'#hashtag *#ArTyouReady*, seguito da *#emptymuseum*. Tuttavia, bisogna notare che questa etichetta è stata utilizzata da una quantità di utenti pari a circa la metà di coloro che hanno creato dei tweet contenenti l'#hashtag *#ArTyouReady*, sebbene i due #hashtag facciano parte della stessa iniziativa e siano stati lanciati da un'unica istituzione in una sola data. Tale differenza può essere dovuta al fatto che *#ArTyouReady* abbia raggiunto una notorietà maggiore e sia stato, per questo motivo, associato a più contenuti, in maniera indipendente da *#emptymusuem*.

**Le istituzioni coinvolte.** La Tabella 4 riporta per ogni #hashtag il numero delle istituzioni che li hanno utilizzati. Dopo aver estratto automaticamente tutti gli utenti per ogni #hashtag, questi sono stati analizzati per essere suddivisi in i) utenti semplici e ii) istituzioni culturali. Di queste due categorie è stata presa in esame la seconda che conta un totale di 389 istituzioni che hanno utilizzato almeno uno dei sei #hashtag. Il gruppo delle istituzioni è stato successivamente suddiviso in base alla tipologia di istituzione culturale, di cui sono state individuate sedici categorie diverse, ridotte a tredici attraverso un ulteriore raggruppamento di tre di queste (Ministero per i beni e le attività culturali e il turismo - MiBACT; Soprintendenza Archeologica Belle Arti e Paesaggio - SABAP; Organizzazione delle Nazioni Unite per l'educazione, la scienza e la cultura - UNESCO) sotto la definizione più ampia di Ente. Tra quelle indicate, si comprende che la categoria Museo è stata quella che ha risposto con maggior prontezza all'uso di tali #hashtag nei momenti di comunicazione social, utilizzando tutti gli #hashtag che presentano un alto numero di interazioni. Da notare che l'#hashtag *#laculturaincasa* è stato utilizzato soltanto da musei, teatri e biblioteche, mentre *#emptymusem*, *#laculturanonsiferma* e *#museichiusimuseiaperti* sono gli #hashtag che presentano una maggiore diffusione tra le tipologie differenti di istituzioni (Tabella 4).

I grafici (Figure 6-10) mostrano nel dettaglio l'uso degli #hashtag da parte di ciascuna istituzione per l'intero periodo considerato. Per alcuni di questi (i.e., *#artyouready* e *#museitaliani*) il MiBACT risulta essere l'utilizzatore più attivo. Tra le istituzioni locali che ricorrono all'hashtag *#museichiusimuseiaperti* per lanciare le proprie iniziative, di particolare rilievo è la presenza del Museo Tattile di Varese, in quanto istituzione creatrice di questo #hashtag e realtà museale, che, sebbene di piccole dimensioni, ha sfruttato il nuovo canale di comunicazione in modo ampio con numerosi tweet. È da sottolineare che, tra le istituzioni che utilizzano gli #hashtag analizzati, il numero dei cosiddetti grandi attrattori è ridotto; notiamo, ad esempio, la presenza del Museo Archeologico di Napoli (MANN) in Figura 7 per i tweet associati a #*emptymuseum*. Il fenomeno della limitata presenza può essere spiegato dal fatto che molti di questi attrattori utilizzavano già nel periodo pre-pandemico degli #hashtag, spesso legati al loro nome e alla loro identità istituzionale molto forte, ed hanno mantenuto tali etichette anche nel periodo oggetto di analisi. In altri casi, invece, nonostante l'associazione di uno degli #hashtag selezionati ai loro contenuti, queste istituzioni hanno privilegiato altre piattaforme social per la veicolazione dei loro contenuti e la comunicazioni con il loro pubblico.

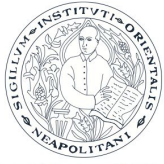
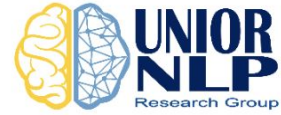

Per ognuno degli account delle istituzioni che hanno partecipato alle iniziative è stata raccolta, laddove presente, la posizione geografica relativa all'account stesso. In seguito, attraverso le librerie di Python geopy[19] e basemap[20], queste posizioni sono state trasformate in coordinate e mappate in relazione al numero di tweet creati dagli account aventi nel testo uno o più degli #hashtag selezionati. I dati raccolti sono presentati in Figura 11, che mostra una corposa concentrazione nelle aree del nord e del centro Italia, dove si ritrovano iniziative più partecipate per gli #hastag selezionati, con un'attività nelle zone del meridione e nelle isole caratterizzata da iniziative che, seppur abbastanza numerose, non possono essere considerate particolarmente rilevanti, ad eccezione di poche, nei centri che hanno istituzioni di un certo rilievo nazionale (ad esempio il Museo Archeologico Nazionale di Napoli - MANN). È da rilevare l'assenza di iniziative da parte delle istituzioni che si trovano nelle regioni dell'Abruzzo, del Molise e della Basilicata.

**Contenuto degli #hashtag e dei tweet.** Dal punto di vista dei contenuti di #hashtag e tweet, è importante precisare che i dati, sia per il primo che per il secondo dataset, sono stati raccolti senza alcuna distinzione sulla lingua utilizzata per il testo del tweet. Una prima analisi, quindi, ha investigato quali lingue sono state maggiormente utilizzate come valore assoluto rispetto a tutti i tweet presi in esame e in relazione ai singoli #hashtag[21]. Ovviamente, trattandosi di iniziative promosse da istituzioni italiane, rivolte principalmente ad un pubblico locale, veicolate attraverso aggregatori in lingua italiana, l'italiano risulta la lingua con un numero maggiore di tweet (Annex - Figura 12). Altre lingue utilizzate risultano l'inglese, seguite dalle lingue francese, catalano, rumeno e spagnolo.

In Figura 13 è riportato il numero di tweet in relazione agli #hashtag specifici e le lingue d'uso per ognuno di questi: si nota una maggioranza dell'uso della lingua italiana per tutti e sei gli

---

[19] https://github.com/geopy/geopy
[20] https://github.com/matplotlib/basemap
[21] La libreria Python language-detection (https://github.com/shuyo/language-detection) è stata utilizzata per estrarre e raccogliere i dati riguardanti le lingue in cui sono stati scritti i tweet contenenti gli #hashtag di interesse.

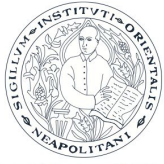 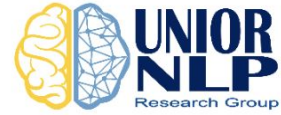

#hashtag, con una sensibile presenza delle etichette *#emptymuseum* e *#ArTyouReady* nei tweet creati da utenti anglofoni. La partecipazione da parte di utenti non italofoni ai contenuti aggregati da questi due #hashtag può essere motivata dal fatto che gli aggregatori sono in lingua inglese, caratteristica che aumenta la possibilità di accesso di un pubblico più ampio e variegato e la conseguente maggiore diffusione degli #hashtag stessi e dei loro contenuti. Inoltre, il numero di tweet non in lingua italiana prodotti utilizzando tali etichette può essere considerato indice della risonanza sovranazionale che hanno avuto le iniziative ad esse collegate. In minore concentrazione, seppur presente, la lingua inglese si registra anche nell'uso degli altri #hashtag propriamente italiani: *#museitaliani*, *#museichiusimuseiaperti*, *#laculturaincasa*, *#laculturanonsiferma*.

Un altro livello di analisi ha riguardato il lessico utilizzato nella creazione degli #hashtag stessi. Per investigare gli aspetti lessicali delle etichette è stato preso in considerazione tutto il set di 33 #hashtag selezionati inizialmente, indipendentemente dal numero di interazioni che hanno catalizzato. Tra questi è possibile notare che le parole scelte per comporre l'#hashtag stesso sono riconducibili a quattro macro-gruppi semantici che fanno riferimento a: i) l'evento a cui sono associate, cioè la quarantena; ii) il luogo di fruizione mutato, cioè non più il museo ma la propria abitazione; iii) il contenuto veicolato, ovvero la cultura; iv) l'iniziativa proposta, come un quiz o una sfida.

Al primo gruppo sono ascrivibili *#quarantinelife*, *#betweenartandquarantine*, *#cultureinquarantine*, che, come si può vedere, sono costruiti sul concetto e in relazione alla condizione di quarantena. In particolare *#cultureinquarantine*, che nel periodo di riferimento è stato il più cliccato, nasce da un'iniziativa promossa dal canale BBC con l'obiettivo di portare nelle case dei fruitori ogni forma d'arte, da quella dei musei a quella del teatro[22].

L'idea di un quasi "delivery" dell'arte è espressa anche dagli #hashtag appartenenti al secondo e al terzo gruppo. Nel secondo gruppo rientrano le etichette che includono un riferimento al luogo di fruizione tradizionale dei contenuti culturali, ovvero il museo, mutato nella sua collocazione geografica, confinato all'interno delle mura domestiche ma che assolve ugualmente alle sue funzioni. Del resto, l'obiettivo del Museo Archeologico Nazionale di Aquileia, che ha lanciato l'iniziativa *#ilMUSEOaCASA*, da cui poi gli #hashtag *#ilmuseoacasa* e *#ilmuseoacasatua*, è stato proprio quello di fare in modo che gli utenti potessero continuare a interagire con i contenuti museali, attraverso una serie di video e attività, suddivisi in tredici episodi pubblicati settimanalmente, rivolte soprattutto ai bambini[23].

Nel terzo gruppo, ugualmente si fa riferimento al mutato luogo di fruizione, ma in questo caso si punta sui contenuti che vengono fruiti in tali luoghi e sul concetto astratto di cultura. Infatti, se l'immagine di un museo a casa può sembrare ingombrante o associabile a quella di "casa museo", che è ben altra cosa, è più facile comprendere l'idea della *#culturaincasa*, come quella che è stata pensata dal Sistema Museale di Roma Capitale, denominato "Musei in Comune" (MIC), che con questo #hashtag propongono, tra le tante iniziative, un tour virtuale in cinque siti archeologici (Musei capitolini, Museo Ara pacis, Museo dei Fori imperiali, Museo Napoleonico, Museo di Villa Torlonia)[24], e moltissime attività indirizzate a bambini e ragazzi, sponsorizzate con l'#hashtag *#laculturaincasaKIDS*.

---

[22] https://www.bbc.co.uk/programmes/articles/45gMYKsDtlRx1WdTCZh3q2b/culture-in-quarantine-tv-radio-and-digital-schedule
[23] https://www.museoarcheologicoaquileia.beniculturali.it/it/287/ilmuseoacasa-iorestoacasa

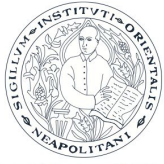 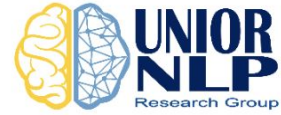

Nell'ultimo gruppo rientrano quegli aggregatori che vogliono proporre iniziative di vario genere e il cui riferimento principale è appunto alle attività proposte/comunicate: da quelle ludiche, come *#museumgames* e *#museumquiz*, a cui si è accennato nella sezione precedente, rivolti a un pubblico ampio e variegato, non solo a quello di giovane età, fino a quelle propriamente riflessive, significanti, cioè, di campagne rivolte a sostegno del settore culturale e dei suoi attori gravemente colpiti dal lockdown, dal punto di vista economico. Tra questi rientra *#ShareOurHeritage*, lanciato dall'UNESCO il 10 aprile 2020, a cui ha fatto seguito il tenace *#ResiliArt* annunciato con un comunicato stampa il 28 aprile 2020[25].

Per quanto riguarda l'estensione del testo dei tweet cui gli #hashtag sono stati associati, la Tabella 5 riporta la lunghezza media dei tweet per ognuno dei sei #hashtag considerati. Tale media è stata ottenuta conteggiando i caratteri, in ottemperanza alle norme di Twitter che permette di formulare testi brevi sulla base di un limite massimo di caratteri e non di parole. È da notare che in media si è fatto ricorso a tutto il numero di caratteri disponibili per i tweet, tranne per gli #hashtag *#emptymuseum* e *#ArTyouReady* che sono ben al di sotto di questa media. Questa differenza è spiegabile sulla base del tipo di iniziative che sono state proposte attraverso questi #hashtag che erano principalmente mirate a richiedere la condivisione di immagini da parte degli utenti; i tweet corredati dal primo #hashtag richiedevano, infatti, la partecipazione del pubblico con le loro foto dei luoghi di cultura senza utenti, quindi, appunto, vuoti, e senza la necessità di inserire un testo descrittivo.

Il contenuto testuale dei tweet è stato analizzato in base alla frequenza dei termini utilizzati nel testo in associazione agli #hashtag selezionati. Per ognuno di questi è stata creata una nuvola rappresentativa del contenuto sulla base del numero di occorrenze delle parole (Figura 14). Rispetto a questi dati, possiamo notare la presenza per tutti i tweet di altri #hashtag di accompagnamento a quello di riferimento. Tali etichette aggiuntive sono di diverso tipo: in alcuni casi si fa riferimento ad #hashtag più generali riferiti alla pandemia (e.g., *litaliachiamo*[26] nei tweet relativi a *#museichiusimuseiaperti* e *coronavirus*, la cui presenza si nota anche in altre nuvole, nei tweet aggregati da *#museitaliani*), in altri il riferimento è ai beni e alle istituzioni culturali più generali (e.g., *mibact* presente sia nella forma di #hashtag che di tag dell'utente _*mibact* nei tweet legati ad #*artyouready* ed #*emptymuseum*). Altri fenomeni interessanti che possono essere osservati in Figura 14 riguardano la presenza di termini che fanno riferimento ai contenuti ed eventi veicolati e aggregati, come nel caso di #*laculturaincasa* in cui compaiono *video*, *concerto, teatro*.

---

[24] http://www.museiincomuneroma.it/it/mostra-evento/la-cultura-casa
[25] https://www.ansa.it/sito/notizie/cultura/arte/2020/04/19/nasce-unesco-resiliart-a-sostegno-della-cultura_90885043-eeb6-4393-89d0-5cbf49203841.html
[26] Questo #hashtag che richiama l'ultima strofa dell'inno nazionale italiano è stato creato a partire dal titolo di una manifestazione online promossa dal MiBACT e svoltasi il 13/03/2020 per raccontare come il paese stava reagendo alla pandemia e alle misure restrittive. Per ulteriori informazioni si veda https://www.beniculturali.it/mibac/export/MiBAC/sito-MiBAC/Contenuti/MibacUnif/Comunicati/visualizza_asset.html_189872393.html

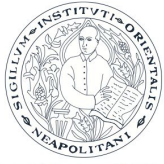
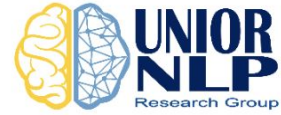

# 4. Conclusioni

In questo report sono stati presentati i risultati iniziali di una indagine sull'uso e la diffusione degli #hashtag da parte delle istituzioni culturali italiane, durante la Fase 1, quella con restrizioni maggiori alla mobilità dei cittadini, del periodo di lockdown dovuto all'epidemia di COVID-19 nel 2020. L'analisi quantitativa ha mostrato dei dati interessanti in termini sia di comunicazione dalle istituzioni verso il pubblico sia delle risposte e dell'interazione degli utenti.

Il fenomeno del moltiplicarsi di #hashtag e tweet, a corredo delle numerose iniziative intraprese in risposta all'evento pandemico, è indicativo della velocità e della proattività con cui le istituzioni, in particolare i musei, hanno reagito. Tale atteggiamento da parte delle istituzioni culturali, come emerge anche dalla prima indagine NEMO, condotta dopo tre settimane dall'inizio del periodo di lockdown, ha contribuito a ridurre lo stato di isolamento e solitudine aumentando i servizi digitali per incoraggiare le persone a rispettare la politica di restrizioni. Inoltre, queste iniziative hanno stimolato un senso di fiducia e comunità attraverso le richieste rivolte agli utenti per la condivisione di oggetti, storie e immagini al fine di creare una memoria collettiva di tale periodo. Infine, le istituzioni hanno dimostrato di poter sostenere il ruolo di educatori informali pure operando a distanza attraverso la proposta di quiz, giochi e materiali educativi fruibili online.

Dal canto loro, gli utenti hanno risposto in maniera positiva a tali iniziative, incrementando drasticamente il numero di interazioni su e con le istituzioni e dimostrando uno spiccato interesse ad una partecipazione attiva nella creazione e trasmissione di contenuti culturali.

Gli #hashtag, scelti come caso di studio per questa analisi, hanno permesso di ricostruire gran parte delle iniziative che le istituzioni hanno lanciato a livello mondiale e nazionale; di monitorarne il successo nel corso dei mesi di marzo, aprile e per i primi giorni di maggio; di individuare tra gli utenti che hanno risposto all'invito a partecipare le istituzioni più attive sui social network e, infine, di essere classificati per aree tematiche a seconda della loro morfologia.

Ulteriori linee di indagine si prospettano a partire dalla prima analisi esplorativa, in particolare per quanto riguarda la possibilità di analisi linguistiche volte a stabilire il tipo di contenuti veicolati e l'atteggiamento degli utenti, in termini di *sentiment* e *sensitivity* ai beni culturali, rispetto alle iniziative e alle informazioni fornite dalle istituzioni in questo periodo.

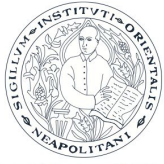
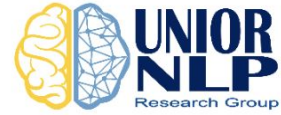

# Bibliografia

# Annex

Tabella 1 - #hashtag relativi alle iniziative delle istituzioni culturali (in grigio quelli esistenti prima della pandemia)

| Hashtag | # Retweet |
| --- | --- |
| #artyouready | 20142 |
| #artchallenge[27] | 16282 |
| #emptymuseum | 12966 |
| #cultureinquaratine | 11888 |
| #museitaliani | 10505 |
| #museichiusimuseiaperti | 6470 |
| #laculturanonsiferma | 3502 |
| #laculturaincasa | 2566 |

---

[27] Come detto precedentemente, l'#hashtag #artchallenge non è stato incluso nel dataset perché si presenta come un #hashtag generico e non emblematico delle iniziative oggetto della presente analisi.

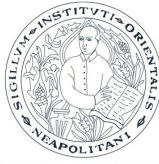
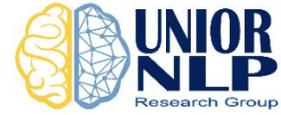

| | |
|---|---|
| #shareourheritage | 2305 |
| #resiliart | 1845 |
| #museumfromhome | 1833 |
| #lartetisomiglia | 1540 |
| #museumalphabet | 1469 |
| #museumfromhome | 1183 |
| #museumgames | 1171 |
| #laculturaincasaKIDS | 599 |
| #laculturacura | 540 |
| #ilmuseoacasa | 379 |
| #ilmuseoacasatua | 316 |
| #museumathome | 202 |
| #museumquiz | 191 |
| #ITweetMuseums | 180 |
| #digitalflashmob | 115 |
| #lartenonsiferma | 115 |
| #nomicoseneimusei | 111 |
| #quarantinelife | 97 |
| #betweenartandquarantine | 89 |
| #ABCBarberiniCorsini | 44 |
| #museiitaliani | 26 |
| #condividilacultura | 21 |
| #stayathomechallenge | 14 |
| #tussenkunstenquarantine | 10 |
| #iogiocodacasa | 6 |

Tabella 2 - Numero di tweet/retweet per i sei #hashtag considerati nel periodo 8/03 -04/05/2020

| Istituzione promotrice | Data lancio | #Hashtag | Tweet | Retweet | Totale |
|---|---|---|---|---|---|
| MiBACT | 29/03[28] | #artyouready | 5310 | 20142 | 25452 |

---

[28] Le istituzioni iniziano a pubblicarlo già il 27/03 per pubblicizzare l'iniziativa, ma la vera data di lancio è il 29/03.

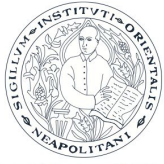
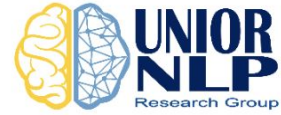

| | | | | | |
|---|---|---|---|---|---|
| MiBACT | 29/03[29] | #emptymuseum | 3204 | 12966 | 16170 |
| Direzione generale Musei - MiBACT | Anteriore al 2020 | #museitaliani | 1874 | 10505 | 12379 |
| Museo Tattile di Varese | 24/02[30] | #museichiusimuseiaperti | 2639 | 6470 | 9109 |
| ICOM Italia | 07/03[31] | #laculturanonsiferma | 1732 | 3502 | 5234 |
| Musei In Comune (MIC) | 17/03 | #laculturaincasa | 1229 | 2566 | 3795 |
| **Totale** | | | **15988** | **56151** | **72139** |

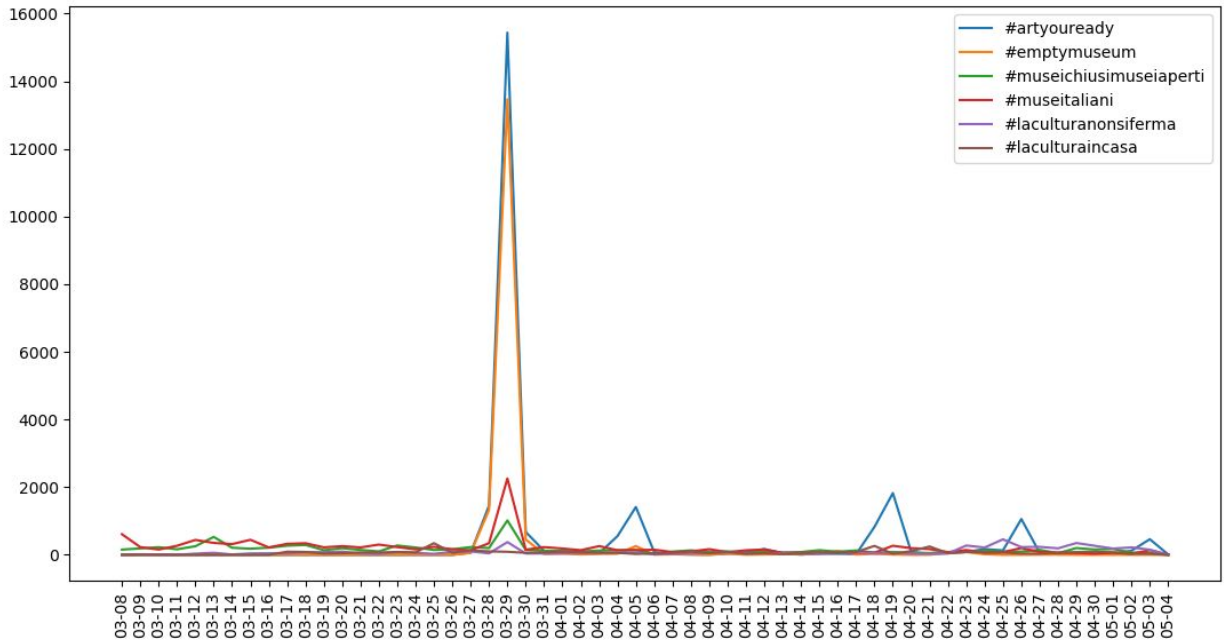

Figura 1 - Andamento del numero di interazioni associati agli #hashtag per l'intero periodo di lockdown

---

[29] Nasce contemporaneamente all' #hashtag #ArTyouReady.
[30] La data di lancio, che è al di fuori del periodo di lockdown che è stato preso in considerazione in questo report, si spiega con la decisione della regione Lombardia di chiudere musei, biblioteche e altri luoghi pubblici di aggregazione, in anticipo rispetto al resto d'Italia, per contenere la diffusione del virus che ha colpito quella zona prima di diffondersi in tutta la Penisola
https://www.varesenoi.it/2020/02/23/leggi-notizia/argomenti/attualita-17/articolo/coronavirus-chiusure-serali-per-bar-e-locali-ecco-lordinanza-completa-della-regione-lombardia.html
[31] Sebbene la data di lancio preceda di un giorno la pubblicazione del DPCM 8 marzo che sancisce la chiusura dei musei, l'iniziativa rientra nel piano di comunicazione adottato da ICOM Italia a sostegno dei Musei al tempo del COVID-19 http://www.icom-italia.org/eventi/laculturanonsiferma/

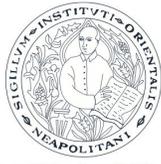
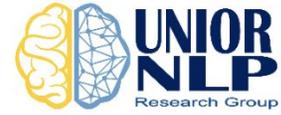

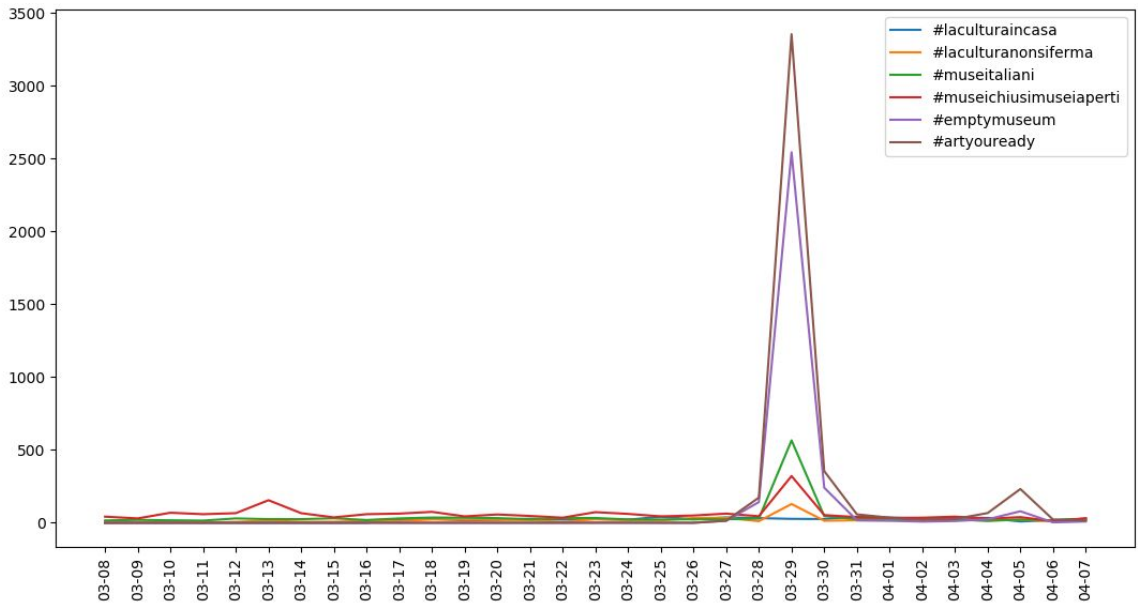

Figura 2 - Andamento degli #hashtag registrato nei tweet formulati nel mese di marzo 2020

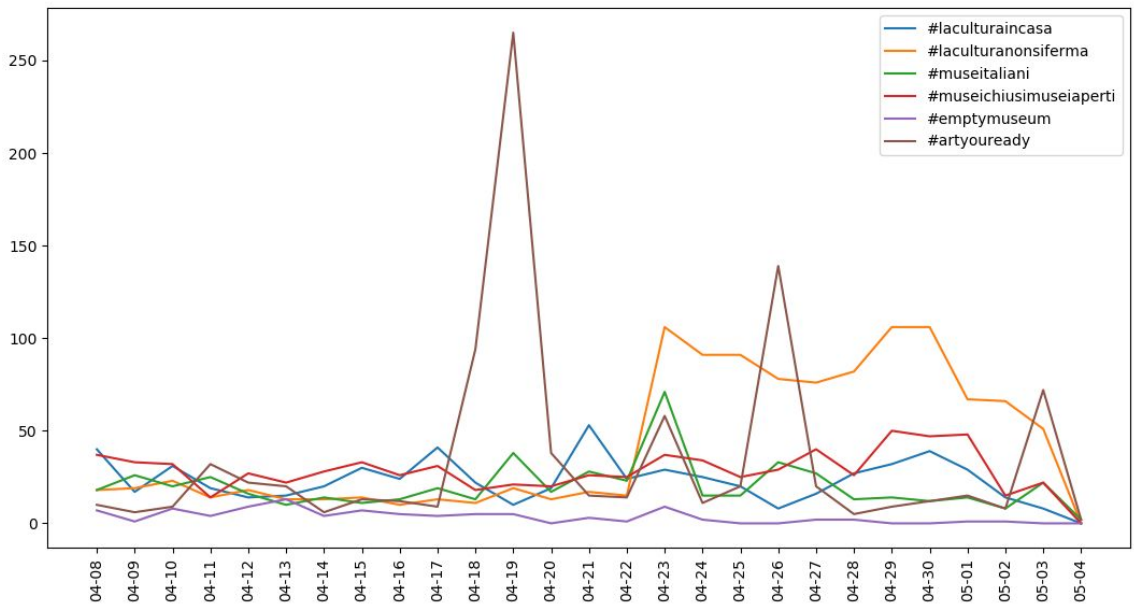

Figura 3 - La figura riporta l'andamento degli #hashtag registrato nei tweet formulati nel mese di aprile e nei primi quattro giorni del mese di maggio 2020

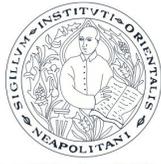
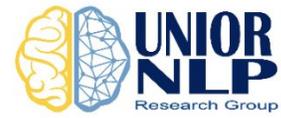

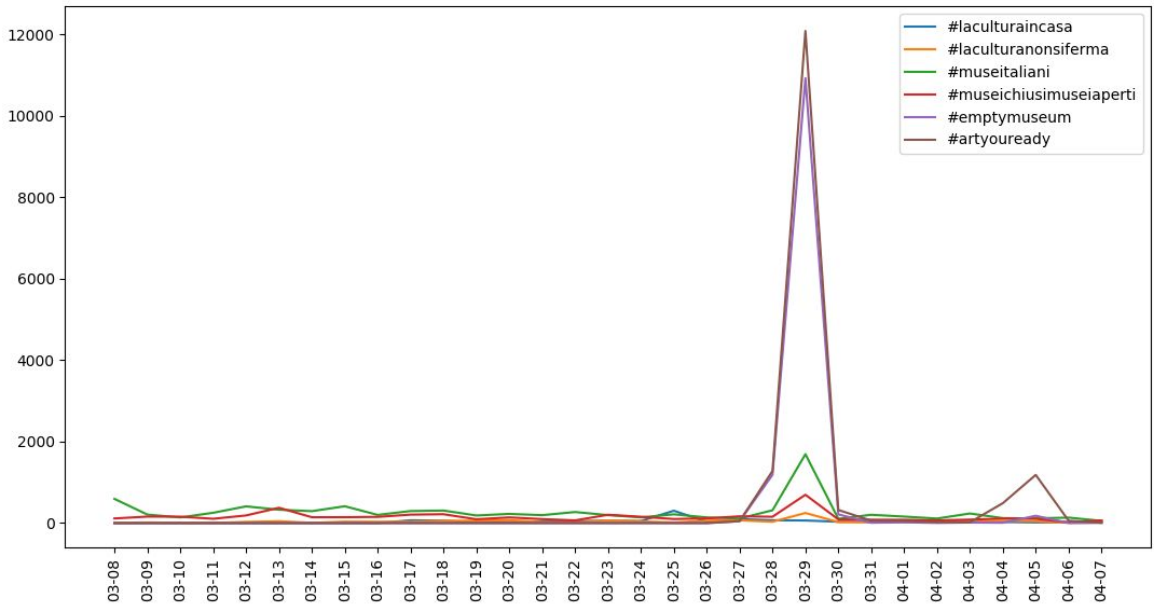

Figura 4 - Andamento degli #hashtag rilevato nei retweet effettuati nel mese di marzo 2020

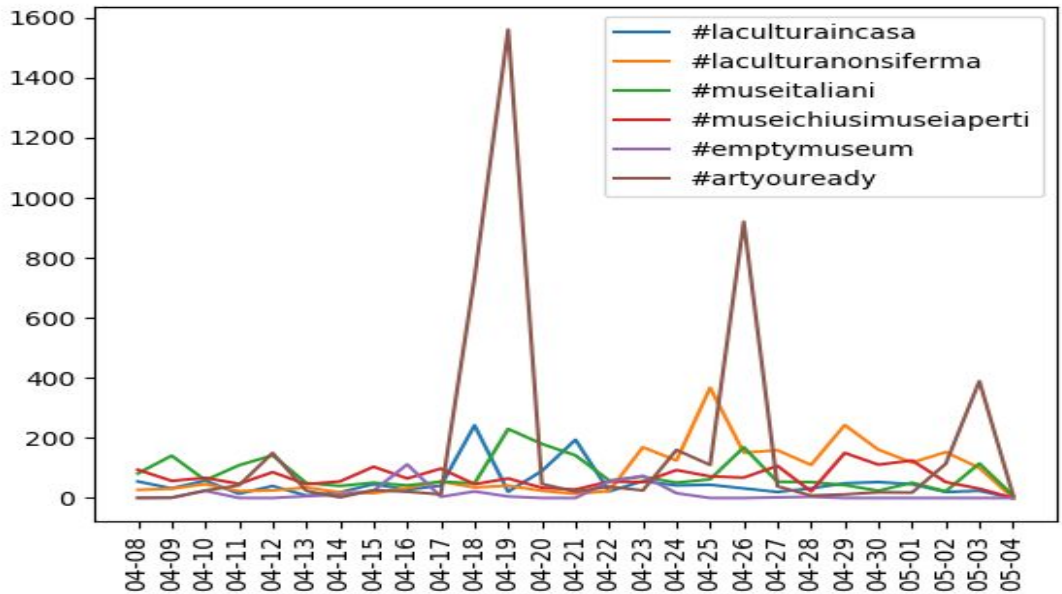

Figura 5 - Andamento degli #hashtag rilevato nei retweet effettuati nel mese di aprile e nei primi quattro giorni del mese di maggio 2020

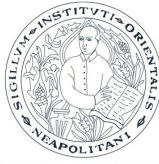
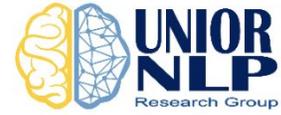

Tabella 3 - Numero di utenti unici per i sei #hashtag considerati nel periodo 8/03 -04/05/2020

| #Hashtag | # Utenti unici |
|---|---|
| #artyouready | 1456 |
| #emptymuseum | 702 |
| #laculturanonsiferma | 414 |
| #museichiusimuseiaperti | 343 |
| #museitaliani | 282 |
| #laculturaincasa | 215 |
| **Totale** | **3412** |
| **Media per #hashtag** | **733.42** |

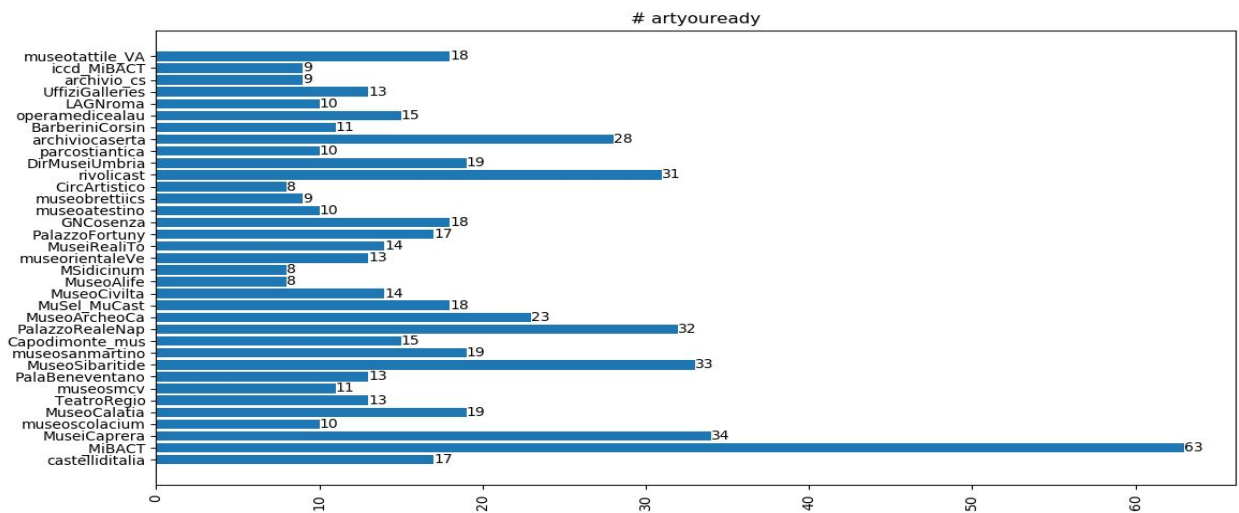

Figura 6 - Numero di tweet lanciati dalle singole istituzioni per l'#hashtag #artyouready

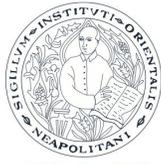
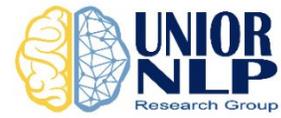
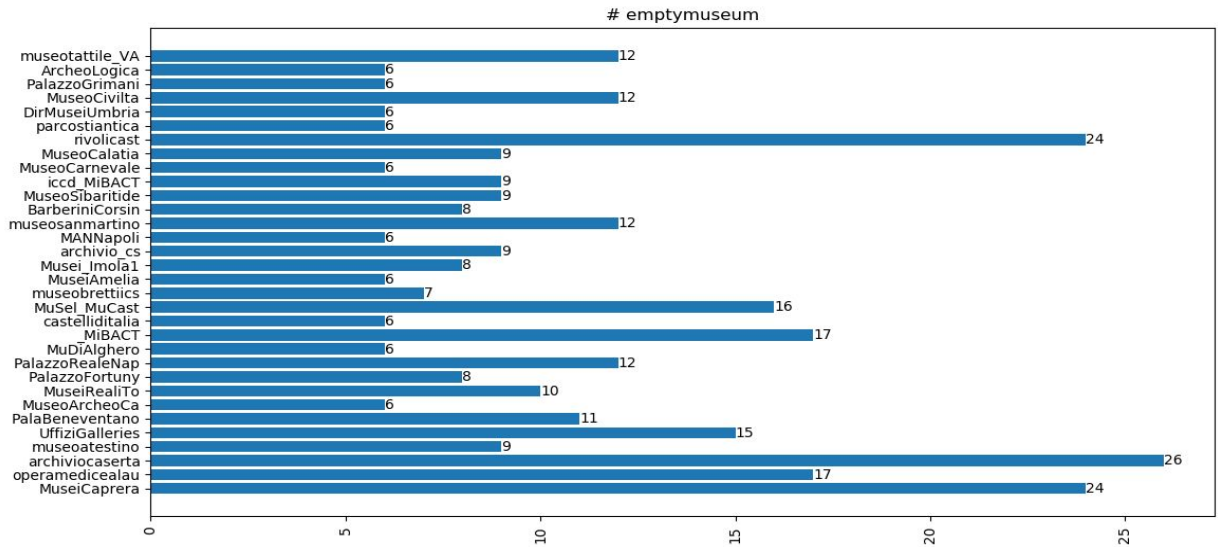

Figura 7 - Numero di tweet lanciati dalle singole istituzioni per l'#hashtag #emptymuseum

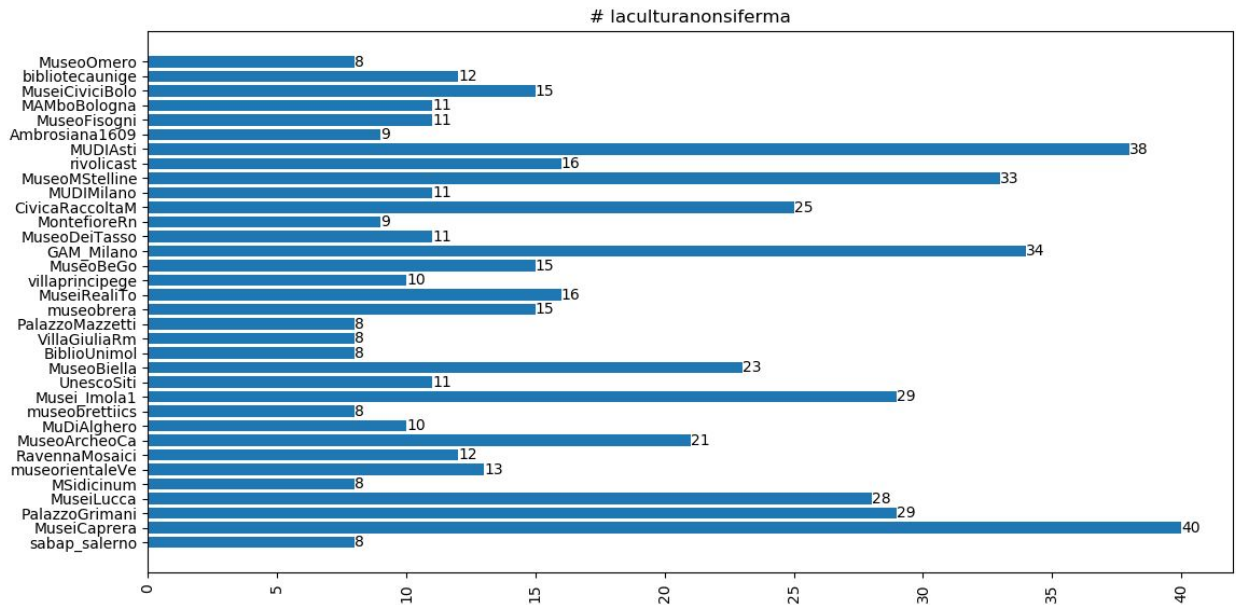

Figura 8 - Numero di tweet lanciati dalle singole istituzioni per l'#hashtag #laculturanonsiferma

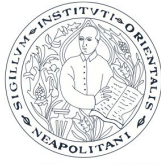
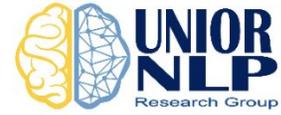
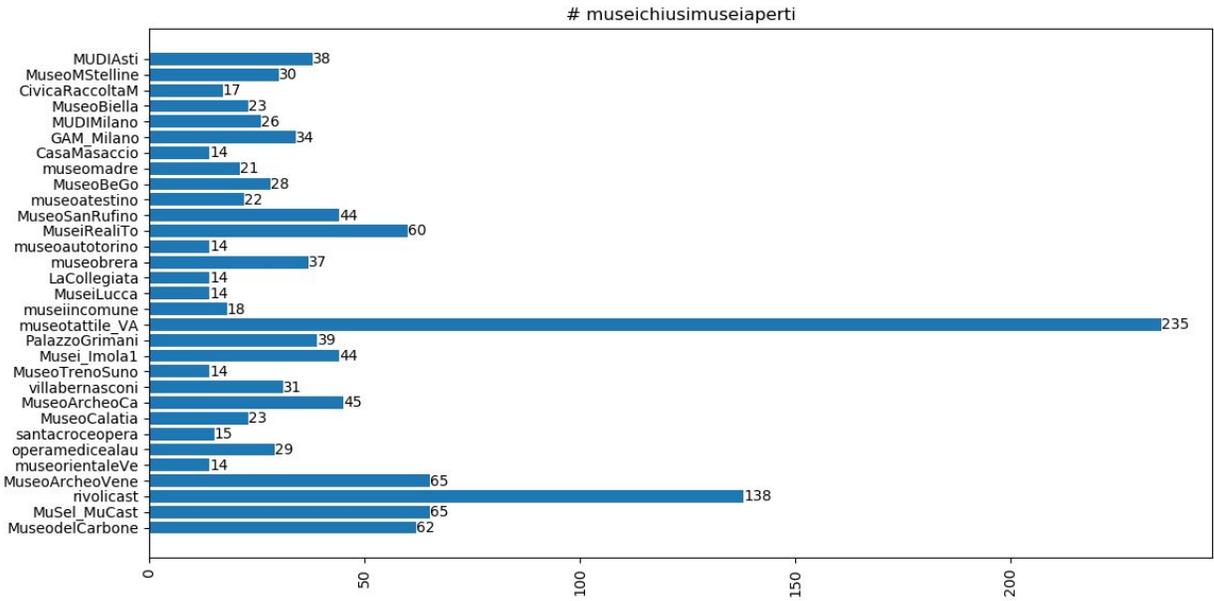

Figura 9 - Numero di tweet lanciati dalle singole istituzioni per l'#hashtag #museichiusimuseiaperti

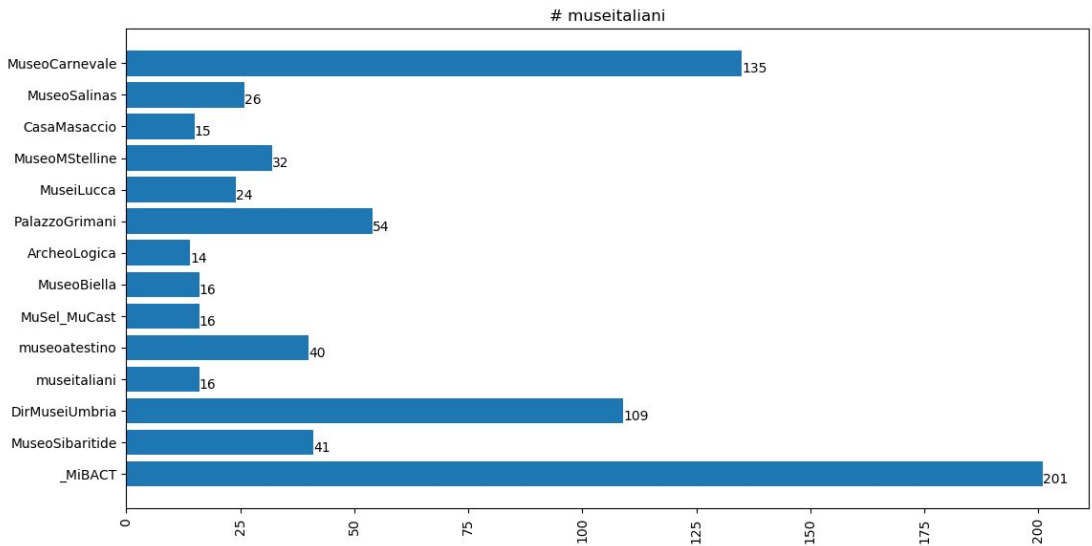

Figura 10 - Numero di tweet lanciati dalle singole istituzioni per l'#hashtag #museitaliani

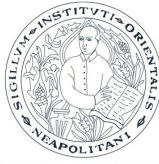

Tabella 4 - La tabella riporta il numero di istituzioni per tipologia che hanno usato i singoli #hashtag

| Istituzione \ Hashtag | #Artyouready | #Emptymuseum | #laculturaincasa | #laculturanonsiferma | #museiItaliani | #museichiusimuseiapertii | Totale |
|---|---|---|---|---|---|---|---|
| **Museo** | 10 | 50 | 4 | 77 | 44 | 93 | **278** |
| **Galleria** | 3 | 6 | N/A | 5 | 3 | 5 | **22** |
| **Biblioteca** | 1 | 3 | 5 | 4 | 2 | 3 | **18** |
| **Palazzo** | N/A | 4 | N/A | N/A | 3 | 5 | **12** |
| **Parco Archeologico** | N/A | 3 | N/A | 2 | 3 | 2 | **10** |
| **Enti** | 1 | 3 | N/A | 3 | N/A | 2 | **9** |
| **Archivio** | N/A | 2 | N/A | 5 | 1 | N/A | **8** |
| **Associazione** | N/A | 1 | N/A | 4 | 2 | 1 | **8** |
| **Teatro** | 1 | N/A | 4 | 3 | N/A | N/A | **8** |
| **Polo museale** | N/A | 2 | N/A | 2 | N/A | 3 | **7** |
| **Castello** | 1 | N/A | N/A | 1 | N/A | 1 | **3** |
| **Reggia** | N/A | 1 | N/A | N/A | 1 | 1 | **3** |
| **Pinacoteca** | N/A | 1 | N/A | 1 | N/A | N/A | **2** |
| **Fondazione** | N/A | 1 | N/A | N/A | N/A | 1 | **1** |
| **Totale istituzioni** | | | | | | | **389** |

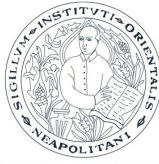
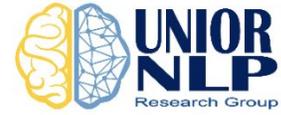

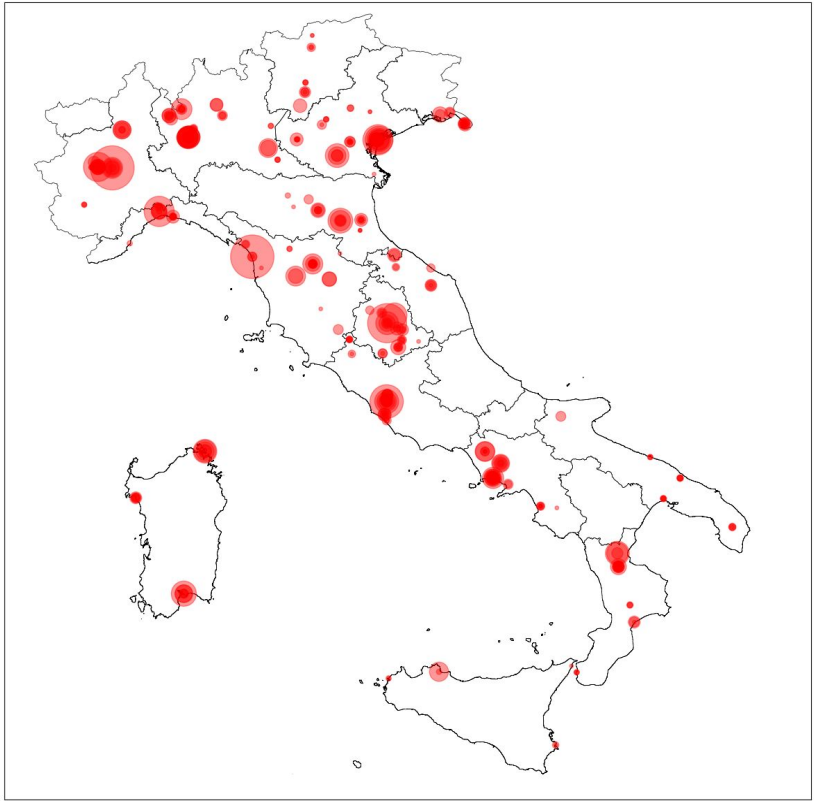

Figura 11 - Distribuzione geografica delle istituzioni che hanno utilizzato almeno uno degli #hashtag selezionati

Tabella 5 - Lunghezza media dei tweet per #hashtag calcolata sulla quantità di caratteri utilizzati

| #Hashtag | Lunghezza media tweet |
|---|---:|
| #laculturanonsiferma | 266.58 |
| #laculturaincasa | 263.87 |
| #museitaliani | 258.29 |
| #museichiusimuseiaperti | 249.49 |
| #emptymuseum | 199.67 |
| #artyouready | 197.38 |
| **Media totale** | **230.86** |

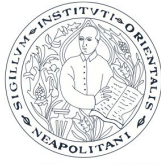
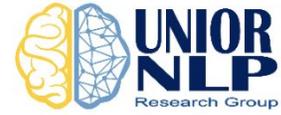

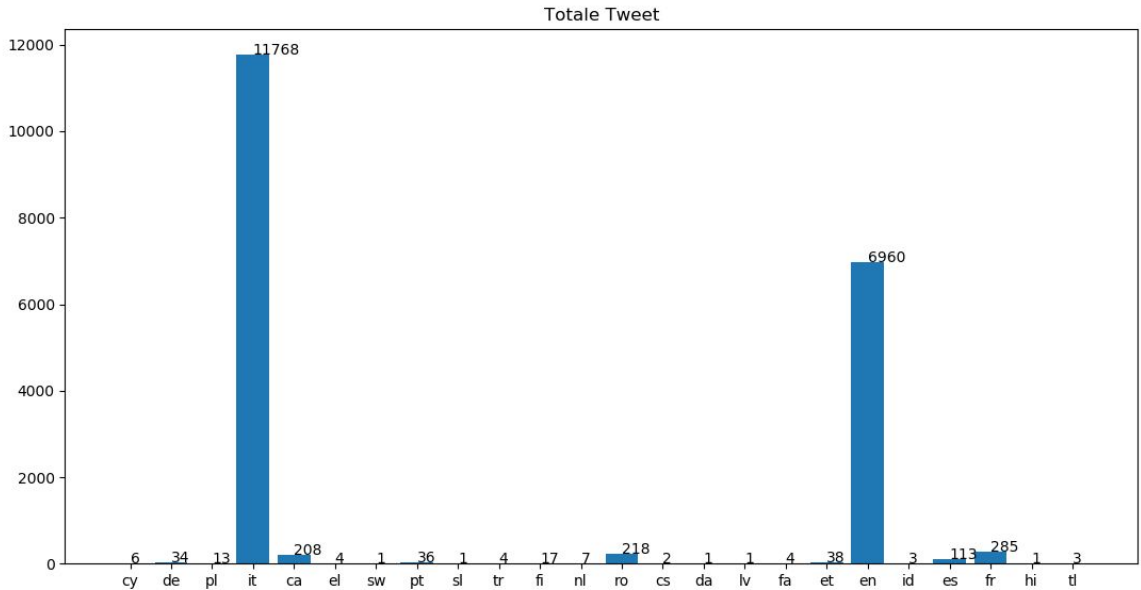

Figura 12 - Numero di tweet per le lingue utilizzate su tutto il dataset

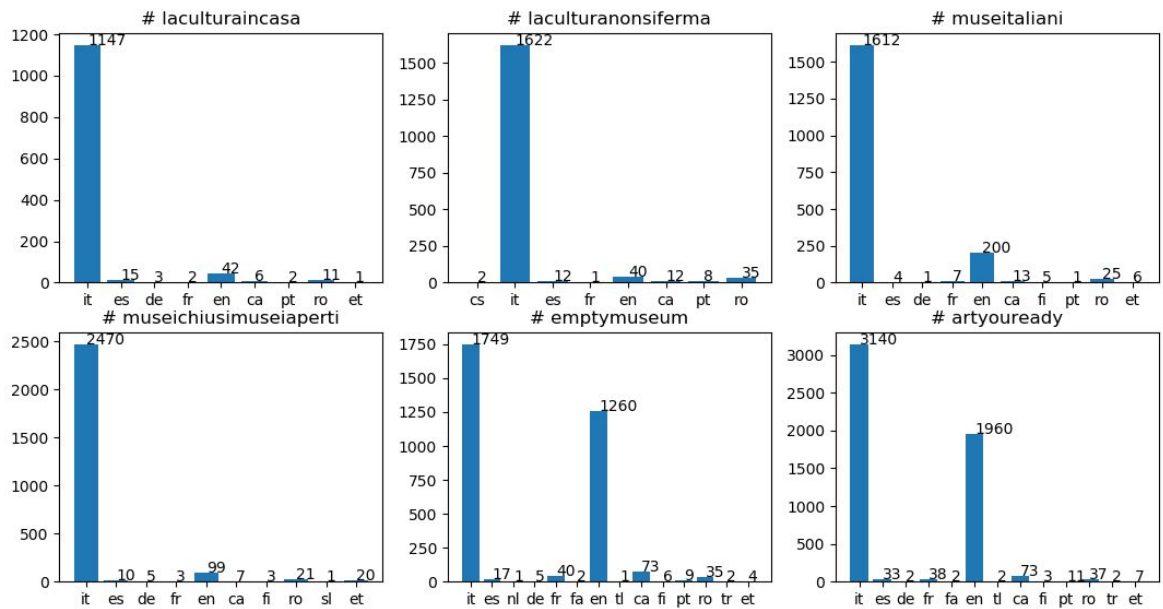

Figura 13 - Numero di tweet per le lingue utilizzate per ognuno dei sei #hashtag

| #ArTyouready | #emptymuseum | #laculturaincasa |
| #laculturanonsiferma | #museichiusimuseiaperti | #museitaliani |

Figura 14 - Nuvole di parole per ciascuno dei sei #hashtag